\documentclass[conference,compsoc]{IEEEtran}

\usepackage{lmodern}
\usepackage{cite}
\usepackage{amsmath}
\usepackage{amssymb}
\usepackage{booktabs}
\usepackage{array}
\usepackage{graphicx}
\usepackage{orcidlink}

\begin{document}

\title{A Dual-Stream Challenge-Response Protocol for Ocular Liveness Verification}

\author{\IEEEauthorblockN{Ismail Kably\,\orcidlink{0009-0009-2323-4399}}
\IEEEauthorblockA{\textit{Konelia Inc.} \\
Austin, Texas, USA \\
\upshape ismail@konelia.com}}

\maketitle

\begin{abstract}
Ocular biometric systems face sophisticated presentation attacks, including high-resolution video replays and real-time generative deepfakes, which easily bypass static liveness checks. Current Presentation Attack Detection (PAD) frameworks typically rely on isolated physiological metrics, such as gaze tracking or the Pupillary Light Reflex (PLR), which can be spoofed independently. This paper proposes a Spatio-Luminance Sensor Fusion protocol, which introduces a dual-stream challenge-response framework for ocular liveness verification by uniting these metrics into a simultaneous authentication challenge. By generating a randomized, time-varying visual stimulus that fluctuates in both spatial trajectory and luminance intensity, we construct a mathematically coupled state-space likelihood model, termed the Synchronization Matrix, to evaluate the continuous cross-correlation between the expected biological latencies of smooth pursuit tracking and pupillary constriction. Using Monte Carlo simulation grounded in literature-derived latency distributions, we demonstrate theoretical separability between genuine and simulated attack conditions, and show that a multi-round challenge design improves the detection of generative deepfakes when a non-zero rendering-latency gap exists. This work provides a simulation-supported theoretical framework for next-generation dynamic spoofing defense in ocular and iris biometrics; human-subject validation is identified as necessary future work before deployment claims can be made.
\end{abstract}


\section{Introduction}
\label{sec:intro}

The proliferation of high-resolution biometric capture devices has escalated the threat of dynamic presentation attacks against ocular recognition systems, which include iris recognition, one of the most widely deployed and well-established biometric modalities \cite{daugman2004}. While traditional spoofing methods relied on static artifacts such as high-resolution photographs or 3D-printed masks, modern threat vectors utilize pre-recorded video replays and real-time generative AI (deepfakes) to bypass standard liveness checks. Deepfake generation has evolved rapidly: early benchmarks focused on GAN-based facial forgery \cite{rossler2019}, while more recent work documents a shift toward diffusion-model-based synthesis capable of producing increasingly convincing facial and ocular detail \cite{bhattacharyya2024}, a trend captured broadly in recent survey and meta-review literature \cite{altuncu2024}.

Existing PAD methodologies typically isolate physiological responses, testing liveness via the Pupillary Light Reflex (PLR) under varying illumination, or via gaze trajectory tracking of an on-screen target. Treating these biological streams independently leaves systems vulnerable: a video replay can spoof a PLR check in isolation, and a sufficiently fast generative model can spoof isolated gaze tracking.

To address this vulnerability, we propose a Spatio-Luminance Sensor Fusion protocol: a simultaneous, dual-stream challenge-response matrix in which a visual target moves along a randomized spatial trajectory while emitting randomized luminance fluctuations. By continuously measuring both smooth pursuit gaze tracking and autonomic pupillary constriction, the system models the eye as a coupled dynamical system. We introduce a Joint Synchronization Metric ($S_{joint}$) to quantify whether spatial movement and pupil dilation are locked in time to the generated challenge, and validate the approach via Monte Carlo simulation against video replay, generative deepfake, and mechanical/prosthetic spoofing.

\textbf{Contributions.} This work makes three main contributions. First, we propose the Spatio-Luminance Sensor Fusion protocol, a dual-stream challenge-response method that jointly drives randomized smooth pursuit and PLR, and define a Joint Synchronization Metric $S_{joint}$ for assessing temporal coupling between gaze and pupil dynamics. Second, we introduce an explicit threat model and latency-based degradation analysis for replay, deepfake, and mechanical/prosthetic attacks against ocular biometrics. Third, we present an in-silico Monte Carlo validation demonstrating theoretical separability under literature-derived latency distributions, and quantify how multi-round challenges improve deepfake detection when a non-zero rendering-latency gap exists.

\section{Related Work and Innovation Gap}
\label{sec:related}

PAD in ocular biometrics has historically relied on isolated, single-stream physiological observations. Broader PAD research spans multiple biometric modalities: Czajka and Bowyer \cite{czajka2018} provide a comprehensive assessment of the state of the art in iris PAD specifically, while Ramachandra and Busch \cite{ramachandra2017} survey PAD methods for face recognition systems more generally, both noting that single-cue liveness signals are increasingly susceptible to sophisticated, AI-driven spoofing. Multimodal biometric fusion --- combining multiple biometric cues to improve robustness and deter spoofing --- has likewise been shown to outperform single-modality systems in general biometric authentication contexts \cite{rossjain2004}, though this principle has not been applied specifically to the fusion of ocular temporal-dynamics signals for PAD, as we propose here.

Considerable research has established the PLR as a liveness indicator. Ellis \cite{ellis1981} demonstrated that pupillary constriction latency is dependent on the intensity of the light stimulus. Bergamin and Kardon \cite{bergamin2003} subsequently established that normal constriction latency falls within 230--357 ms. Standalone PLR frameworks remain susceptible to pre-recorded video replay, since the response to a predictable light pulse can be captured and replayed.

Gaze-tracking mechanisms enforce liveness by requiring users to follow a randomized on-screen target. Makowski et al. \cite{makowski2021} demonstrated deep-learning-based oculomotoric identification combined with presentation-attack detection using eye movement dynamics; however, this and comparable gaze-based liveness approaches rely on spatial and motion cues alone and do not incorporate the pupillary light reflex as a complementary autonomic signal. The visual system initiates smooth pursuit tracking with an open-loop latency of approximately 80--150 ms, depending on target velocity and step-ramp conditions \cite{carlgellman1987, rashbass1961}. While effective against static image spoofing, isolated spatial tracking is increasingly vulnerable to generative deepfakes capable of synthesizing realistic eye movements. Commercial and patent literature on joint pupil--corneal modeling for gaze estimation accuracy treats PLR-induced diameter change as an optical error source rather than a security primitive.

Recent spatio-temporal fusion methods have proven highly effective in adjacent presentation-attack domains: Khan et al. \cite{khan2025} apply spatio-temporal deep learning to face PAD, while Mohamed et al. \cite{mohamed2026} target gait spoofing with a dedicated spatio-temporal network. These results confirm the value of jointly modeling spatial and temporal features for biometric security, yet this paradigm has not been extended to coupled ocular dynamics --- the simultaneous modeling of gaze trajectory and pupillary temporal response. This paper identifies precisely that gap: the absence of a temporally coupled, multi-stream PAD framework. No current methodology fuses a continuous spatial tracking trajectory with simultaneous pupillary luminance response into a single, mathematically synchronized challenge-response matrix architected to defeat dynamic, AI-driven presentation attacks.

\section{The Spatio-Luminance Protocol Architecture}
\label{sec:architecture}

\begin{figure*}[t]
    \centering
    \includegraphics[width=\textwidth]{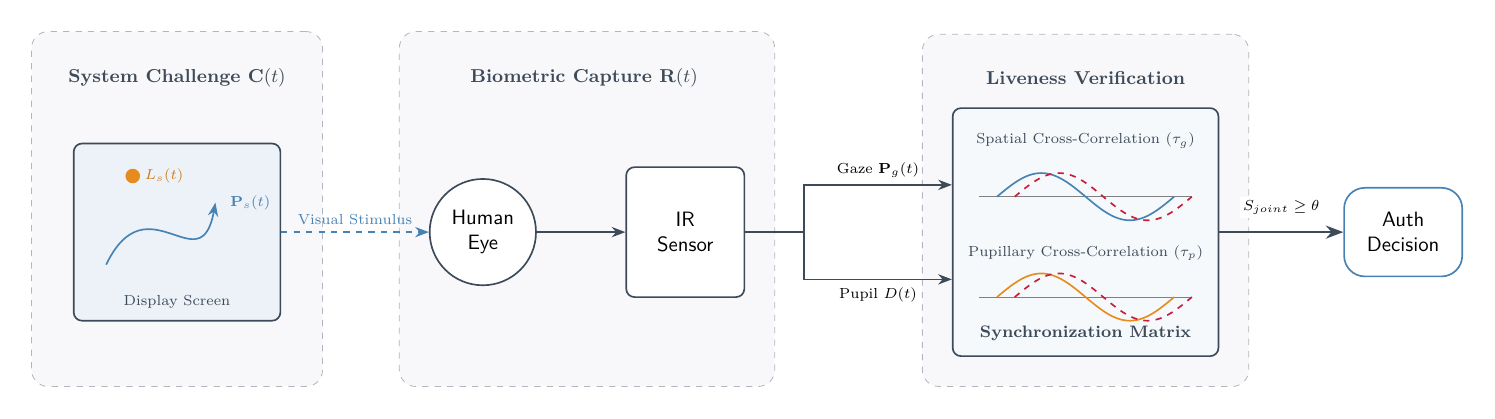}
    \caption{Architecture of the Spatio-Luminance Sensor Fusion protocol. The system issues a randomized spatial and luminance challenge (left), captures the biological smooth pursuit and pupillary responses (center), and evaluates their temporal cross-correlation within the Synchronization Matrix (right).}
    \label{fig:architecture}
\end{figure*}

The proposed protocol comprises three stages, illustrated in Fig.~\ref{fig:architecture}: a challenge generator, a biometric capture front-end, and the synchronization engine.

\subsection{The Spatial Challenge}
\label{subsec:spatial}
The system generates a continuous, unpredictably randomized spatial trajectory $\mathbf{P}_s(t)$ on a graphical display. The path avoids linear geometric patterns and uses randomized directional shifts to prevent predictive saccades and force continuous smooth pursuit correction.

\subsection{The Luminance Challenge}
\label{subsec:luminance}
The system simultaneously modulates the luminance intensity $L_s(t)$ of the moving target, uncoupled from the spatial trajectory, using randomized frequency and amplitude variation.

\subsection{Biological Response Capture}
\label{subsec:capture}
An infrared biometric sensor captures ocular response at high frame rate. A computer vision pipeline extracts two synchronized streams: gaze position $\mathbf{P}_g(t)$ and normalized pupil diameter $D(t)$.

\section{The Spatio-Luminance Synchronization Matrix}
\label{sec:sync}

\subsection{State Variable Definitions}
\label{subsec:state}

The system-generated challenge at time $t$:
\begin{equation}
\mathbf{C}(t) = \begin{bmatrix} \mathbf{P}_s(t) \\ L_s(t) \end{bmatrix} = \begin{bmatrix} x_s(t) \\ y_s(t) \\ L_s(t) \end{bmatrix}
\label{eq:challenge}
\end{equation}

where $\mathbf{P}_s(t) \in \mathbb{R}^2$ is the randomized target path and $L_s(t) \in [L_{\min}, L_{\max}]$ is the randomized target luminance.

The captured biometric response:
\begin{equation}
\mathbf{R}(t) = \begin{bmatrix} \mathbf{P}_g(t) \\ D(t) \end{bmatrix} = \begin{bmatrix} x_g(t) \\ y_g(t) \\ D(t) \end{bmatrix}
\label{eq:response}
\end{equation}

where $\mathbf{P}_g(t)$ is estimated gaze position and $D(t)$ is normalized pupil diameter.

\subsection{Biological Transfer Functions and Latency Modeling}
\label{subsec:transfer}

\textbf{Gaze tracking (smooth pursuit).} Carl and Gellman \cite{carlgellman1987} report latency bounds of approximately 80--100 ms for predictable, high-velocity targets; Rashbass \cite{rashbass1961} reports latencies closer to 150 ms for step-ramp stimuli. Because our protocol uses an unpredictable trajectory forcing continuous re-initiation of pursuit, we adopt the conservative bound and define $\mu_g = 150$ ms as the population-mean baseline used throughout this paper. We model:
\begin{equation}
\mathbf{P}_g(t) = \mathbf{P}_s(t - \tau_g) + \boldsymbol{\eta}_g(t), \quad \boldsymbol{\eta}_g(t) \sim \mathcal{N}(0, \boldsymbol{\Sigma}_g)
\label{eq:gaze}
\end{equation}
where $\boldsymbol{\Sigma}_g$ (parameterized by $\sigma_g$) is an assumed variance pending empirical calibration.

\textbf{Pupillary light reflex.} Bergamin and Kardon \cite{bergamin2003} report a constriction latency window of 230--357 ms. We adopt the window midpoint, $\mu_p = 290$ ms, as the population-mean baseline used throughout this paper. The dynamic response is modeled as:
\begin{equation}
\frac{dD(t)}{dt} = -\frac{1}{\tau_k}[D(t) - D_0] - \gamma L_s(t - \tau_p) + \eta_p(t)
\label{eq:plr_ode}
\end{equation}
where $\tau_k$ is the pupillary recovery time constant, $\gamma$ is the light-sensitivity gain, and $\eta_p(t) \sim \mathcal{N}(0, \sigma_p^2)$ represents pupillary hippus, with $\sigma_p$ an assumed variance pending empirical calibration.

\subsection{The Joint Synchronization Metric}
\label{subsec:metric}

\begin{equation}
\begin{split}
S_{joint} = {} & w_1 \left[\max_{\tau_g} \rho\big(\mathbf{P}_g(t), \mathbf{P}_s(t-\tau_g)\big)\right] \\
& + w_2 \left[\max_{\tau_p} \rho\big(D(t), \hat{D}(t|L_s)\big)\right] \\
& - w_3 \, \Phi(\tau_g, \tau_p)
\end{split}
\label{eq:sjoint}
\end{equation}

where $\rho(A,B)$ is the Pearson correlation coefficient over a sliding window $T_w$, and $\hat{D}(t|L_s)$ is the predicted pupil diameter obtained by integrating the PLR equation using historical luminance data. The biological plausibility penalty is:
\begin{equation}
\begin{split}
\Phi(\tau_g, \tau_p) = {} & \frac{(\tau_g - \mu_g)^2}{2\sigma_g^2} + \frac{(\tau_p - \mu_p)^2}{2\sigma_p^2}, \\
& \mu_g = 150 \text{ ms}, \; \mu_p = 290 \text{ ms}
\end{split}
\label{eq:phi_penalty}
\end{equation}

For preliminary modeling we adopt $w_1 = w_2 = 1.0$, $w_3 = 2.0$. Authentication succeeds if $S_{joint} \geq \theta_{threshold}$. Final values for $w_1, w_2, w_3$, and $\theta_{threshold}$ are scoped as future work, to be derived from a target False Acceptance Rate during empirical calibration (Section~\ref{sec:conclusion}).

\section{In-Silico Validation}
\label{sec:validation}

\begin{figure*}[t]
    \centering
    \includegraphics[width=0.6\textwidth]{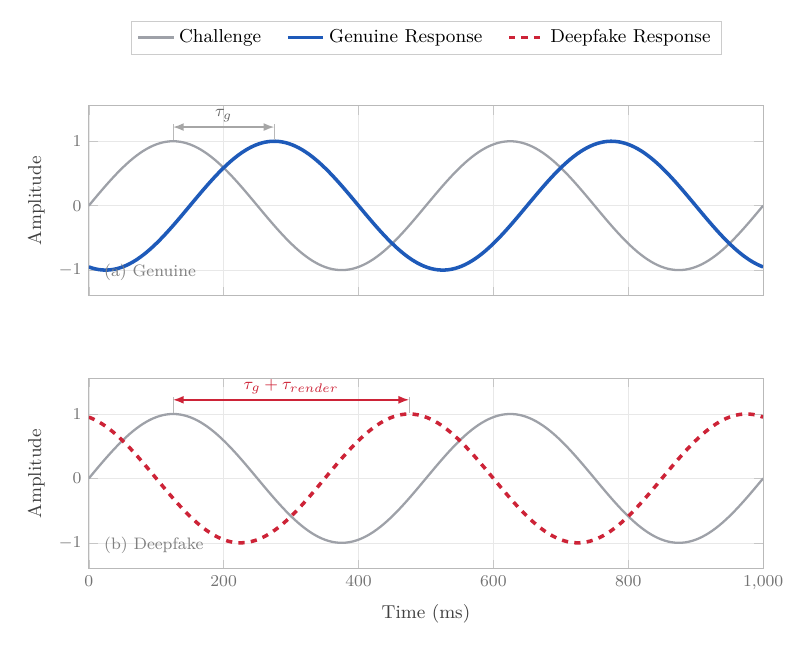}
    \caption{Timing comparison. A genuine response lags the challenge by the biological latency $\tau_g$, whereas a real-time deepfake incurs an additional rendering lag $\tau_{render}$ that shifts its response further out of phase.}
    \label{fig:latency}
\end{figure*}

The temporal divergence between genuine biological latencies and generative artifacts is depicted in Fig.~\ref{fig:latency}, which highlights the rendering-induced phase shift $\tau_{render}$ that our model exploits.

\subsection{Simulation Methodology}
\label{subsec:methodology}

We conducted a Monte Carlo simulation of the Joint Synchronization Metric, $S_{joint}$, using the literature-derived latency distributions established in Section~\ref{sec:sync} ($\mu_g = 150$ ms, $\mu_p = 290$ ms) \cite{bergamin2003, carlgellman1987, rashbass1961} and the weights defined in Section~\ref{subsec:metric} ($w_1 = w_2 = 1.0$, $w_3 = 2.0$). This validation is simulation-based, not derived from human-subject or empirical attack data, and is intended to establish theoretical separability rather than to serve as a performance benchmark.

Four conditions were modeled over $N = 10{,}000$ trials each: genuine response ($\tau_g \sim \mathcal{N}(150, \sigma_g^2)$ ms, $\tau_p \sim \mathcal{N}(290, \sigma_p^2)$ ms); video replay, modeled by drawing $\tau_g, \tau_p$ from $\mathcal{U}(0, 1000)$ ms to represent a response decorrelated by construction from the live randomized challenge; generative deepfake, modeled as the genuine distribution offset by a swept rendering latency $\tau_{render}$; and mechanical/prosthetic spoofing, modeled with near-zero latency but an explicit $5\times$ reduction applied to the pupillary correlation term $\rho_p$ to represent the waveform-shape mismatch between a rigid mechanical aperture and the biological first-order response curve. Correlation terms were approximated via a Gaussian decay function of latency deviation, and $S_{joint}$ was computed per Section~\ref{sec:sync}. Detection was assessed primarily via receiver operating characteristic analysis, reporting area under the curve (AUC) and equal error rate (EER); for the rendering-latency sweep (Section~\ref{subsec:results}), we additionally report detection rate at a fixed operating point (the 95th-percentile threshold of the genuine score distribution) as a complementary, more operationally interpretable statistic computed from the same $S_{joint}$ scores. A sensitivity analysis was conducted over $\sigma_g \in [10, 30]$ ms and $\sigma_p \in [15, 50]$ ms.

We additionally evaluated a multi-round extension in which $M$ independent randomized sub-challenges are issued per session, with the session score taken as the mean $S_{joint}$ across rounds. This models an attacker who must respond independently to each fresh challenge. Detection was compared for $M \in \{1, 3, 5, 10\}$ across the same $\tau_{render}$ sweep, with $\sigma_g$ and $\sigma_p$ held fixed at 20 ms and 30 ms, respectively.

\subsection{Results}
\label{subsec:results}

Replay and mechanical spoofing were strongly separable from genuine responses (AUC = 0.996 and 1.000, respectively). Deepfake detectability depended strongly on rendering latency, with AUC rising from chance (0.502) at $\tau_{render} = 0$ ms to 0.717 at 25 ms, 0.948 at 50 ms, and $\geq 0.997$ beyond 75 ms. Pooled AUC remained above 0.93 across the tested $\sigma_g$ and $\sigma_p$ range, indicating robustness to current parameter uncertainty.

The multi-round extension improved deepfake detection whenever $\tau_{render} > 0$ (Table~\ref{tab:multiround}). At $\tau_{render} = 25$ ms, AUC rose from 0.717 ($M=1$) to 0.966 ($M=10$), and the reliable-detection threshold ($\text{AUC} \geq 0.90$) dropped from 50 ms to 25 ms with $M \geq 5$. At $\tau_{render} = 0$ ms, multi-round averaging provided no benefit; AUC remained at chance for all $M$, as averaging independent samples cannot introduce separability absent an underlying latency gap. We interpret this as a security--latency tradeoff: increasing $M$ narrows the range of rendering speeds that evade detection, at the cost of longer authentication sessions, and should be tuned to anticipated adversary capability.

\begin{table}[t]
\centering
\caption{AUC versus deepfake rendering latency ($\tau_{render}$) and number of challenge rounds ($M$); reliable detection is reached at lower latency as $M$ increases, with no benefit at $\tau_{render}=0$ ms.}
\label{tab:multiround}
\begin{tabular}{rrrrr}
\toprule
$\tau_{render}$ (ms) & $M=1$ & $M=3$ & $M=5$ & $M=10$ \\
\midrule
0   & 0.502 & 0.502 & 0.501 & 0.492 \\
10  & 0.541 & 0.582 & 0.594 & 0.643 \\
25  & 0.717 & 0.843 & 0.902 & 0.966 \\
50  & 0.947 & 0.997 & 1.000 & 1.000 \\
75  & 0.997 & 1.000 & 1.000 & 1.000 \\
100 & 1.000 & 1.000 & 1.000 & 1.000 \\
150 & 1.000 & 1.000 & 1.000 & 1.000 \\
200 & 1.000 & 1.000 & 1.000 & 1.000 \\
\bottomrule
\end{tabular}
\end{table}

\textbf{Limitations.} This analysis assumes i.i.d. rendering latency across rounds. A real adversarial pipeline may exhibit correlated latency across rounds --- for instance, consistently fast or slow relative to its own fixed hardware and model architecture --- in which case the reported gains may not fully generalize, since correlated errors average out less effectively than independent ones. Characterizing this requires adversarial red-teaming with real generative pipelines or a correlated-latency attacker model, alongside the human-subject calibration proposed in Section~\ref{sec:conclusion}.

\section{Threat Model and Advanced Threat Vectors}
\label{sec:threat_model}

\subsection{Attack Scenarios and Mathematical Degradation}
\label{subsec:attacks}

\textbf{Attack 1: Pre-recorded video replay.} A recorded response reacts to a historical challenge $\mathbf{C}_{hist}(t)$, uncorrelated with the live randomized challenge $\mathbf{C}_{live}(t)$. Both correlation terms collapse toward zero, driving $S_{joint} < \theta_{threshold}$. Video replay is a long-studied attack vector in face and ocular biometrics generally, with dedicated benchmark databases and baseline countermeasures established in the literature \cite{chingovska2012}; our contribution is to show that the same decorrelation principle extends naturally to a jointly randomized spatial-luminance challenge.

\textbf{Attack 2: Real-time generative deepfake.} An attacker's rendering pipeline incurs latency $\tau_{render}$, shifting observed latencies to $\tau_g = \mu_g + \tau_{render}$ and $\tau_p = \mu_p + \tau_{render}$. The penalty term $\Phi(\tau_g, \tau_p)$ grows quadratically with this deviation, driving $S_{joint}$ below threshold once $\tau_{render}$ is non-negligible (quantified in Section~\ref{sec:validation}).

\textbf{Attack 3: Actuated mechanical/prosthetic spoof.} A mechanical aperture exhibits linear dynamics lacking the first-order biological recovery curve and spontaneous hippus. The resulting waveform mismatch between $D(t)$ and $\hat{D}(t|L_s)$ suppresses $\rho_p$, preventing authentication despite low latency. This attack class builds on prior work modeling liveness detection against mechanical eye replicas via oculomotor plant characteristics \cite{komogortsev2015}; our contribution extends this principle to the joint spatial-luminance domain.

\subsection{Limitations and Advanced Threat Vectors}
\label{subsec:limitations}

Two limitations bound the framework's guarantees as presented in this paper.

First, deepfake rejection depends on $\tau_{render}$ producing an observed latency outside the biological penalty function's effective tolerance. Section~\ref{sec:validation} quantifies this bound directly: detection is near chance as $\tau_{render} \to 0$ and approaches ceiling only once $\tau_{render}$ exceeds roughly 50--75 ms (or approximately 25 ms under a multi-round challenge design, Section~\ref{subsec:results}). Should hardware-accelerated generative rendering approach zero latency, this framework --- using timing analysis alone --- would not detect the resulting deepfake; this is a stated theoretical floor, not an incidental gap.

Second, this protocol does not inherently defeat a presentation relay (``human-proxy'' or man-in-the-middle) attack, in which a live human proxy views the relayed challenge and their genuine ocular response is captured and relayed to the sensor. Because a real human eye is tracking the target, $S_{joint}$ would evaluate favorably. Defeating this attack class requires pairing the synchronization check with continuous verification of iris texture identity, which is outside the scope of the temporal model presented here.

\section{Conclusion and Future Work}
\label{sec:conclusion}

This paper introduced a Spatio-Luminance Sensor Fusion protocol to secure ocular biometric systems against dynamic presentation attacks, moving beyond single-stream liveness checks toward a simultaneous, multi-dimensional challenge-response framework. We formalized the Synchronization Matrix ($S_{joint}$) and, in Section~\ref{sec:validation}, provided Monte Carlo evidence of theoretical separability between genuine and simulated attack conditions, including a quantified characterization of the framework's dependence on adversarial rendering latency.

Future work required to move this framework toward deployment includes: (1) an empirical human-subject calibration study to determine $\sigma_g$, $\sigma_p$, and to tune $w_1, w_2, w_3$ and $\theta_{threshold}$ against a target False Acceptance Rate; (2) hardware benchmarking of IR sensor and display refresh rates needed to capture $\rho$ at millisecond precision; (3) adversarial red-teaming with real generative rendering pipelines to test the i.i.d. latency assumption underlying the multi-round extension (Section~\ref{subsec:results}); and (4) integration of continuous iris texture verification to mitigate presentation relay attacks (Section~\ref{subsec:limitations}).

In practical terms, the proposed framework is most naturally deployed as an additional liveness layer on top of existing iris or face recognition pipelines, rather than as a standalone biometric, providing timing-based resistance against replay and deepfake attacks once calibrated with real user and adversarial data.


\begin{thebibliography}{16}

\bibitem{daugman2004}
J. Daugman, ``How iris recognition works,'' \emph{IEEE Trans. Circuits Syst. Video Technol.}, vol. 14, no. 1, pp. 21--30, 2004.

\bibitem{rossler2019}
A. R\"ossler, D. Cozzolino, L. Verdoliva, C. Riess, J. Thies, and M. Nie\ss ner, ``FaceForensics++: Learning to detect manipulated facial images,'' in \emph{Proc. IEEE/CVF Int. Conf. Computer Vision (ICCV)}, 2019, pp. 1--11.

\bibitem{bhattacharyya2024}
C. Bhattacharyya, H. Wang, F. Zhang, S. Kim, and X. Zhu, ``Diffusion deepfake,'' arXiv preprint arXiv:2404.01579, 2024.

\bibitem{altuncu2024}
E. Altuncu, V. N. L. Franqueira, and S. Li, ``Deepfake: Definitions, performance metrics and standards, datasets, and a meta-review,'' \emph{Front. Big Data}, vol. 7, art. 1400024, 2024.

\bibitem{czajka2018}
A. Czajka and K. W. Bowyer, ``Presentation attack detection for iris recognition: An assessment of the state-of-the-art,'' \emph{ACM Comput. Surv.}, vol. 51, no. 4, pp. 86:1--86:35, 2018.

\bibitem{ramachandra2017}
R. Ramachandra and C. Busch, ``Presentation attack detection methods for face recognition systems: A comprehensive survey,'' \emph{ACM Comput. Surv.}, vol. 50, no. 1, pp. 8:1--8:37, 2017.

\bibitem{rossjain2004}
A. Ross and A. K. Jain, ``Multimodal biometrics: An overview,'' in \emph{Proc. 12th European Signal Processing Conference (EUSIPCO)}, Vienna, Austria, 2004, pp. 1221--1224.

\bibitem{ellis1981}
C. J. Ellis, ``The pupillary light reflex in normal subjects,'' \emph{Br. J. Ophthalmol.}, vol. 65, no. 11, pp. 754--759, 1981.

\bibitem{bergamin2003}
O. Bergamin and R. H. Kardon, ``Latency of the pupil light reflex: Sample rate, stimulus intensity, and variation in normal subjects,'' \emph{Invest. Ophthalmol. Vis. Sci.}, vol. 44, no. 4, pp. 1546--1554, 2003.

\bibitem{makowski2021}
S. Makowski, P. Prasse, D. R. Reich, D. Krakowczyk, L. A. J\"ager, and T. Scheffer, ``DeepEyedentificationLive: Oculomotoric biometric identification and presentation-attack detection using deep neural networks,'' \emph{IEEE Trans. Biom. Behav. Identity Sci.}, vol. 3, no. 4, pp. 506--518, 2021.

\bibitem{carlgellman1987}
J. R. Carl and R. S. Gellman, ``Human smooth pursuit: Stimulus-dependent responses,'' \emph{J. Neurophysiol.}, vol. 57, no. 5, pp. 1446--1463, 1987.

\bibitem{rashbass1961}
C. Rashbass, ``The relationship between saccadic and smooth tracking eye movements,'' \emph{J. Physiol.}, vol. 159, no. 2, pp. 326--338, 1961.

\bibitem{khan2025}
S. Khan, T. H. M. Siddique, M. S. Ibrahim, A. J. Siddiqui, and K. Huang, ``Spatio-temporal deep learning for improved face presentation attack detection,'' \emph{Knowl.-Based Syst.}, 2025, art.\ no.\ 113059.

\bibitem{mohamed2026}
I. Mohamed, A. Salah, E. Debie, M. Abdellah, and A. Abdellatif, ``GaitSpoofNet,'' \emph{Front. Artif. Intell.}, vol. 9, 2026, art.\ no.\ 1821341.

\bibitem{chingovska2012}
I. Chingovska, A. Anjos, and S. Marcel, ``On the effectiveness of local binary patterns in face anti-spoofing,'' in \emph{Proc. Int. Conf. Biometrics Special Interest Group (BIOSIG)}, 2012, pp. 1--7.

\bibitem{komogortsev2015}
O. V. Komogortsev, A. Karpov, and C. D. Holland, ``Attack of mechanical replicas: Liveness detection with eye movements,'' \emph{IEEE Trans. Inf. Forensics Security}, vol. 10, no. 4, pp. 716--725, 2015.

\end{thebibliography}
\end{document}